\documentclass{article}

\usepackage{times}  
\usepackage{helvet}  
\usepackage{courier}  
\usepackage{url}  
\usepackage{graphicx}  
\usepackage{mathtools}
\usepackage{amsmath,amssymb,amsfonts}

\usepackage{caption}
\usepackage{amsthm,dsfont}
\newcommand{\ignore}[1]{}
\usepackage{color}
\usepackage{authblk}
\usepackage{algorithm}
\usepackage{algorithmic}
\usepackage{graphicx}
\usepackage{float} 
\usepackage{multirow}
\usepackage{amsfonts}
\usepackage{comment}
\usepackage{fancyhdr}
\usepackage{rotating}
\newtheorem{theorem}             {Theorem}

\begin{document}
	
	\title{Specific Single- and Multi-Objective Evolutionary Algorithms for the Chance-Constrained Knapsack Problem}

\author{Yue Xie}
\author{Aneta Neumann}
\author{Frank Neumann}
\affil{Optimisation and Logistics, School of Computer Science,\\ The University of Adelaide, Adelaide, Australia}
\renewcommand\Authands{ and }
\maketitle
\begin{abstract}
The chance-constrained knapsack problem is a variant of the classical knapsack problem where each item has a weight distribution instead of a deterministic weight.
The objective is to maximize the total profit of the selected items under the condition that the weight of the selected items only exceeds the given weight bound with a small probability of $\alpha$. In this paper, consider problem-specific single-objective and multi-objective approaches for the problem. We examine the use of heavy-tail mutations and introduce a problem-specific crossover operator to deal with the chance-constrained knapsack problem. Empirical results for single-objective evolutionary algorithms show the effectiveness of our operators compared to the use of classical operators. Moreover, we introduce a new effective multi-objective model for the chance-constrained knapsack problem. We use this model in combination with the problem-specific crossover operator in multi-objective evolutionary algorithms to solve the problem. Our experimental results show that this leads to significant performance improvements when using the approach in evolutionary multi-objective algorithms such as GSEMO and NSGA-II.
\end{abstract}

\section{Introduction}

Evolutionary Algorithms (EAs) are bio-inspired randomised optimisation techniques and have been widely applied to many stochastic combinatorial optimisation problems \cite{HORNG20123603,TILL2007630,RAKSHIT201718,6148271}. In this paper, we study the chance-constrained knapsack problem (CCKP), which is a stochastic version of the classical knapsack problem, where each item has a random weight based on a known distribution and independent of the weights of other items. The goal of the chance-constrained knapsack problem is to select a subset of items with the maximum profit when satisfying the chance constraint. The chance constraint is satisfied if the probability that the weight of the selected subset violates the knapsack capacity if at most $\alpha$, where $\alpha$ is a given parameter. Chance constraint optimisation problems have so far received little attention in the evolutionary computation literature \cite{DBLP:journals/tec/LiuZFG13} although they capture many relevant stochastic real-world settings. Doerr et al. \cite{DBLP:journals/corr/abs-1911-11451} investigated submodular optimisation problems with chance constraints and analysed the approximation behaviour of greedy algorithms. 
Recently, Xie et al. \cite{Xie} first applied the evolutionary algorithms in solving the CCKP. So far, the chance-constrained knapsack problem has not received much attention in the evolutionary computation literature, and the goal of this paper is to pursue this critical research direction further.

Evolutionary algorithms (EAs) have been applied to many combinatorial optimisation problems and proven to be very successful in solving complex optimisation problems \cite{HORNG20123603,6148271,RAKSHIT201718,shi2017reoptimization,chiong2012variants}. Mutation operators and crossover operators are the core features of evolutionary algorithms that have been studied by many researchers in the last decades \cite{beasley1993overview1,beasley1993overview2}. The operators are used to guide the algorithm towards a solution to a given problem, and they play different roles to improve the solutions produced by the algorithm. The mutation operators are used to maintaining the diversity of the solution space, and the crossover operators combine the current chromosomes of solutions into new solutions \cite{mitchell1998introduction}. The operators can succeed in finding a good solution when dealing with combinatorial optimisation problems.

However, previous studies of the stochastic knapsack problems have not dealt with discussing the performance of evolutionary algorithms in solving the problems. There are several versions of the stochastic knapsack problems that have been studied in the literature~\cite{kosuch2010upper,merzifonluouglu2012static}. These studies aim to maximize the expected profit resulting from the assignment of items to the knapsack. Some researchers consider approximation algorithms for stochastic knapsack problems~\cite{Anand,Dean,DRISCOLL1996265}. Recently, for CCKP, Goyal and Ravi \cite{goyal2010ptas} presented a fully polynomial-time approximation scheme (FPTAS) for the case when item sizes are normally distributed while satisfying the chance-constrained strictly. Klopfenstein and Nace \cite{klopfenstein2008robust} designed a pseudo-polynomial time resolution algorithm for the chance-constrained knapsack problem and provided to obtain feasible solutions. Han et al. \cite{han2016robust} proved that the resulting robust knapsack problem with the polyhedral uncertainty set could be solved by solving ordinary knapsack problems repeatedly, which makes it then possible to solve the problem in pseudo-polynomial time. Assimi et al. \cite{DBLP:journals/corr/abs-2002-06766} studied the dynamic chance-constrained knapsack problem and proposed another objective function to deal with the dynamic capacity of the knapsack. 

In this work, we consider the same chance-constrained knapsack problem that has already been examined by Xie et al.~\cite{Xie}. In their study, the authors have applied Chebyshev's inequality and Chernoff bound to estimate the probability of constraint violation. They have reformulated the chance-constrained knapsack problem as a multi-objective model concerning the total profit and the probability of chance-constrained. To improve the algorithms presented in this paper, we introduce a problem-specific crossover operator and examine the use of heavy-tail mutation operator in dealing with the chance-constrained knapsack problem. Moreover, we apply the operators in single-objective and multi-objective evolutionary algorithms to solve the CCKP. The problem-specific crossover operator is a combination of uniform crossover operator and greedy method.  A uniform crossover operator allows the offspring chromosomes to search all possibilities of re-combining the different genes in parents \cite{syswerda1989uniform,page1999mutation,falkenauer1999worth,CHICANO201476}. The heavy-tailed mutation operators have been regarded in many sub-fields of evolutionary computation \cite{771163,Yao}, and proved to be effective in solving the combinatorial optimization problems \cite{Benjamin,Friedrich}. Besides, we introduce a new effective multi-objective model for the chance-constrained knapsack problem, which improves diversity in the solution space according to the problem.

The remaining parts of the paper are organized as follows. In the next section, we describe the chance-constrained knapsack problem. In Section \ref{operators}, we introduce the heavy-tail mutation operator and the new problem-specific crossover operator. Section \ref{singlesection} presents the single-objective evolutionary approach for the chance-constrained knapsack problem and describes experimental results. In Section \ref{multiapp}, we introduce a new multi-objective model for the chance-constrained knapsack problem and present empirical results. Section \ref{conclusion} concludes the paper.

\section{Preliminaries}
\label{pre}

\subsection{Problem formulation}
Let $N=\{1,...,n\}$ be a set of items with vectors $p \in \mathbb{R}_{+}^n$ and $w \in \mathbb{R}_{+}^n$ assigning positive profits and weights to the items. In addition, a knapsack capacity $C>0$ is given. The classical knapsack problem can be defined as: $\max_{x \in \{0,1\}^n} \{\sum_{i \in N}p_i x_i \mid \sum_{i \in N}w_i x_i \leq C\}$. Hence, the goal is to find a selection of items that have maximum profit among all sets of items that do not violate the capacity of knapsack. We consider the search space is $\{0,1\}^n$ and a candidate solution $x=\{x_1,...,x_n\} \in \{0,1\}^n$ is a bit-string of length $n$, where item $i$ is chosen iff $x_i =1$. 

In the chance-constrained knapsack problem, we assume that the weight vector $w$ is not known with accuracy, $w$ can take on values according to a given probability distribution. In this paper, we assume that the weights of the items are independent of each other. In the chance-constrained knapsack problem, the goal is to maximize the profit of the selection items under the condition that the probability of violating the knapsack constraint is less than a given threshold of $\alpha$. Formally, the chance-constrained knapsack problem is given as:
\begin{align}
	\mathbf{Maximize} \quad &P(x)=\sum_{i \in N} p_ix_i\\
	& Prob(\sum_{i \in N }w_i x_i \geq C) \leq \alpha
	\label{chance} \\
	 & x\in \{0,1\}^n
\end{align}
where $\alpha \in [0,1]$ is a small value. We are looking for a solution $x$ of maximum profit that violates the capacity bound $C$ with probability at most $\alpha$.

\subsection{Surrogate Functions Based on Tail Bounds}
\label{surrogate}
In this section, we introduce some surrogate functions presented in \cite{Xie} which are constructed by well-known deviation inequalities, namely Chebyshev's inequality and Chernoff bound, to tackle the chance constraints (\ref{chance}) by evolutionary algorithms. 

Chebyshev's inequality has a high utility because it can be applied to any probability distribution with known expectation and standard deviation of design variables. It also gives a tighter bound in comparison to the weaker tails such as Markov's inequality \cite{doerr2018probabilistic}. The standard Chebyshev's inequality is two-side and provides tails for upper and lower bounds \cite{casella2002statistical}. As we are only interested in the probability of violating the knapsack capacity $C$, we use a one-sided Chebyshev's inequality known as Cantelli's inequality \cite{DasGupta2008}. For brevity, we refer to this one-sided Chebyshev's inequality as Chebyshev's inequality in this paper.
 \begin{theorem}[Chebyshev's inequality]
 \label{thm:cheb}
   Let $X$ be a random variable with expectation $\mu_X$ and standard deviation $\sigma_X$. Then for any $k\in \mathbb{R}^+$, 
   \begin{displaymath}
     Prob(X\geq\mu_X +k)\leq \frac{\sigma_X^2}{\sigma_X^2+k^2}.
   \end{displaymath}
 \end{theorem}
We assume that the weights of the items are all independent to each other and that the weight of items have uniform distributions and take values in a real interval $[a_i-\delta,a_i + \delta]$, where $a_i$ is the expected weight of item $i$ and $\delta$ is a parameter that determines the uncertainty of the weights. 

We use the surrogate functions given in \cite{Xie} to transform the constraint (\ref{chance}) by applying Chebyshev's inequality are as follow:
\begin{align}
     Prob( W\geq C)\leq\frac{\delta^2\sum_{i=1}^n x_i}{\delta^2\sum_{i=1}^n x_i+3(C- \sum_{i=1}^n a_i x_i)^2}.
\end{align}

Compared to Chebyshev's inequality, Chernoff bounds provides a sharper tail with exponential decay behaviour. In Chernoff bounds, the random variable is a summation of independent random variables, which take on values in $[0,1]$ \cite{motwani_raghavan_1995}. There are several types of Chernoff bounds. In this paper, we use the following one which proposed by \cite{doerr2018probabilistic} (Theorem 10.1).
 \begin{theorem}[Chernoff bound]
 \label{thm:cher}
   Let $X_1,...,X_n$ be independent random variables taking values in $[0,1]$. Let $X=\sum_{i=1}^n X_i$. Let $\epsilon\geq 0$. Then
   \begin{equation}
   Prob(X\geq(1+\epsilon)E(X)) \leq \left(\frac{e^\epsilon}{(1+\epsilon)^{(1+\epsilon)}}\right)^{E(X)}.
   \label{func:cher}
   \end{equation}
 \end{theorem}
Only in the case that the weight of the items is independent to each other and have an additive uniform distribution, we can apply the Chernoff bound to the chance constraint, then we have the surrogate function as follow:
\begin{equation}
  \label{chernoff:fun}
   Prob(W\geq C) 
\leq \left(\frac{e^{\frac{C-E_W(X)}{\delta \sum_{i=1}^n x_i}}}{\left(\frac{\delta \sum_{i=1}^n x_i +C-E_W(X)}{\delta\sum_{i=1}^n x_i}\right)^{\frac{\delta \sum_{i=1}^n x_i +C-E_W(X)}{\delta\sum_{i=1}^s x_i}}}\right)^{\frac{1}{2}\sum_{i=1}^n x_i}
  \end{equation}
Here $W=\sum_{i \in I} w_i x_i$ and $E_W(X)=\sum_{i \in I} a_i x_i$ are the summary weight and summary expected weight of selection items. For details on how to obtain this inequality, we refer to Theorem 3.3 in~\cite{Xie}. 

Considering the surrogate functions of the chance constraint, we distinguish two types of infeasible solutions as we consider small values of $\alpha$, i.e. $\alpha=10^{-3}$. The first type of infeasible solutions have expected weight that exceeds the capacity of the knapsack, the probability of violating the constraint is at least $1/2$ due to our assumptions on the probability distributions. In the second type of infeasible solutions, the expected weight is less than the capacity, but the probability of violating the capacity is larger than the given $\alpha$.  

\section{SPECIFIC OPERATORS FOR CCKP}
\label{operators}

In this section, we propose a mutation operator and a crossover operator to work in evolutionary algorithms when solving the chance-constrained knapsack problem. In terms of mutation, we propose to use heavy-tail mutations that have recently gained significant attention, particularly in the area of theory of evolutionary computation~\cite{Benjamin,Friedrich}.

Compare to the standard mutation operator, the heavy-tail mutation operator can flip more than one bit in each step, and it has shown to be useful in some single-objective combinatorial optimization problems~\cite{Friedrich}. Therefore, we examine the use of heavy-tail mutation operator in single-objective evolutionary algorithms to deal with the chance-constrained knapsack problem. The proposed problem-specific crossover operator (PS crossover operator) combines uniform crossover and a greedy method, and generates an offspring to inherit the common "genes" from the parents selects more effective uncommon "genes". 

\subsection{Heavy-Tail Mutation Operator}

Doerr et al.~\cite{Benjamin} pointed out that when a multi-bit flip is necessary to leave a local optimum, it needs much time to find the right bits to be flipped if using standard bit mutations. Then higher mutation probabilities may be justified. Neumann and Sutton~\cite{fogaruntime} proved that even for the most straightforward cases of CCKP, it is possible to have local optima in the search space that are difficult to escape when using standard bit mutations. The conclusion of \cite{fogaruntime} motivates our investigations on applying heavy-tail mutations for CCKP.

The heavy-tail mutation operator overcomes the mentioned negative effect when using standard bit mutations, and is at the same time structurally close to the traditional way of performing mutations. There is a general belief that a dynamic choice of the mutation rate as done in heavy-tail mutation can be profitable. Theoretical studies show that the performance of the $(1+1)$~EA using a heavy-tail mutation operator is better than the standard $(1+1)$~EA in solving \textit{jump} functions \cite{Benjamin}. 

In the heavy-tail mutation operator, the mutation rate is chosen randomly in each iteration according to a power-law distribution with (negative) exponent $\beta>1$. The heavy-tailed choice of the mutation rate ensures that with probability $\circleddash(k^{-\beta})$, and exactly $k$ bits are flipped. 
The power low distribution is given as follows.
\begin{theorem}[Discrete power-law distribution: $D_{n/2}^{\beta}$]
Let $\beta>1$ be a constant. If a random variable $X$ follows the distribution $D_{n/2}^{\beta}$, then 
\begin{align}
    Prob({X=\theta}) = \left(C_{n/2}^{\beta}\right)^{-1} \theta^{-\beta}
\end{align}
for all $\theta \in [1,..,n/2]$, where the normalization constant is $C_{n/2}^{\beta}:=\sum_{i=1}^{n/2}i^{-\beta}$.
\label{power-law}
\end{theorem}

In this paper, we use the definition of the heavy-tail mutation operator proposed in~\cite{Benjamin} as follow:
when the parent individual is a bit string $x\in\{0,1\}^n$, the mutation operator first chooses a random mutation rate $\theta/n$ with $\theta\in [1,..,n/2]$ chosen according to the power-law distribution $D_{n/2}^{\beta}$ and then creates an offspring by flipping each bit of string independently with probability $\theta/n$. The working principle of this operator is given in Algorithm \ref{heavyTail}.

\begin{algorithm}[t]
\caption{The heavy-tail mutation operator}
\begin{algorithmic}[1]
\STATE $x=\{x_1,..,x_n\} \in \{0,1\}^n$;
\STATE Choose $\theta\in [1,..,n/2]$ randomly according to $D_{n/2}^{\beta}$;
\FOR{$j=1$ to $n$}
\IF{$rand([0,1]) \leq \theta/n$} 
\STATE $y_i \leftarrow 1-x_i$
\ELSE  
\STATE $y_i \leftarrow x_i$
\ENDIF
\ENDFOR
\STATE return $y=\{y_1,..,y_n\}$
\label{heavyTail}
\end{algorithmic}
\end{algorithm}

\subsection{Problem Specific Crossover Operator}

The proposed crossover operator is a combination of the uniform crossover and considers the specificity of CCKP and the standard KP. The uniform crossover operator can easily preserve all parent similarities when generation new offspring. Indeed, for many combinatorial optimization problems, good solutions being close in the objective space are expected to be rather similar in the decision space \cite{jaszkiewicz2001multiple,ishibuchi2008empirical}. Therefore, the uniform crossover can maintain the so-called good gene combinations which are constructed during the search process. 

The problem-specific crossover operator, which we shall call PS crossover operator, adapts the benefit of the uniform crossover operator. For all genes that are different in the two parents, we evaluate the quality of these genes, specifically to KP. We use the \textit{profit/weight} ratio to determine the quality of genes. Then, genes (items) are sorted in descending order according to the quality, and we apply a greedy insertion heuristic to iterative inserts a candidate item that has the highest profit/weight ratio. In this stage, we insert the first $k$ items according to the ordering of the non-common genes, where $k$ is a number that randomly chooses according to the Normal distribution $k\sim N\left(\frac{m}{2},\frac{m}{2}\right)$, $m$ denotes the number of genes where the two parents differ. 

\section{Single-Objective Approaches}
\label{singlesection}
In this section, we first introduce the single-objective evolutionary algorithm that we study in this paper. Then we examine the impact of the heavy-tail mutation operator in the algorithm. Furthermore, we test the performance of a population-base evolutionary algorithm using the problem-specific crossover operator for solving the chance-constrained knapsack problem.

\subsection{Evolutionary Algorithms}
\label{oneplusone}
The first single-objective evolutionary algorithm we consider is $(1+1)$~EA, which is the most simple evolutionary algorithm. $(1+1)$~EA is also known as a baseline single-objective optimization algorithm to solve the chance-constrained knapsack problem in the previous research \cite{Xie}. 

\begin{algorithm}[t]
\caption{$(1+1)$~EA}
\begin{algorithmic}[1]
\STATE Choose $x\in \{0,1\}^n$ uniformly at random.
\WHILE { \textit{stopping criterion not meet}}
\STATE $y\leftarrow$ flip each bit of $x$ independently with probability of $p$;
\IF{$f(y)\geq f(x)$} 
\STATE $x \leftarrow y$ ;
\ENDIF
\ENDWHILE 
\end{algorithmic}
\label{1+1ea}
\end{algorithm}

The $(1+1)$~EA, given in Algorithm \ref{1+1ea}, initializes a random solution $x\in\{0,1\}^n$. In the main optimization loop, one offspring $y$ is generated by flipping each bit of the parent with probability $p$. The offspring replaces the parent unless it has an inferior fitness. In this paper, we define the fitness of a solution $x$ same to the one in the study of Xie et al. \cite{Xie}:

\begin{eqnarray}
f(X)=(u(x), v(x), P(x))
\label{fitsingle}
\end{eqnarray}

where $u(x)=max\{\sum_{i \in N}a_i x_i -C,0\}$, $v(x)=max\{Prob(\sum_{i \in N} w_i x_i \geq C)-\alpha,0\}$, $P(x)=\sum_{i \in N} p_i x_i$. For this fitness function, $u(x)$ and $v(x)$ need to be minimized and $P(x)$ maximized respectively, and it is optimized in the lexicographic order. Usually, in the mutation step, the probability of flip each bit is $1/n$, where $n$ is the length of a solution. In this paper, we take this as a standard mutation probability and compare it with the heavy-tail mutation operator.

We then introduce a population-based single-objective evolutionary algorithm to deal with the chance-constrained knapsack problem instances. This kind of algorithm maintains a population of binary solutions presented as a bit string. We set the population size to $10$ and examine the performance of this $(\mu+1)$~EA 
using the heavy-tail mutation operator and the problem-specific crossover operator separately and its combination. Algorithm~\ref{10+1ea} is the $(\mu+1)$~EA using the heavy-tail mutation operator and the PS crossover operator.

\begin{algorithm}[t]
\caption{$(\mu+1)$~EA}
\begin{algorithmic}[1]
\STATE Randomly generate $\mu$ initial solutions as the initially population; 
\WHILE {\textit{stopping criterion not meet}}
\STATE Choose $x_1\in \{0,1\}^n$ and $x_2\in \{0,1\}^n$ uniformly at random from the population $X$; $x_1 \ne x_2$.
\STATE apply the PS crossover operator in $x_1$ and $x_2$, generate an offspring $y$;
\STATE apply the heavy-tail mutation operator to $y$;
\IF{$y$ is better than the worse solution in $X$}
\STATE replace the worst solution with $y$;
\ENDIF
\ENDWHILE 
\label{10+1ea}
\end{algorithmic}
\end{algorithm}

\subsection{Experimental Setup}
\label{expset}
The benchmarks used in this paper are the same as in~\cite{Xie}. We consider two types of instances: \textit{Uncorrelated} and \textit{Bounded Strong Correlated}, for each instance, the weights of items have uniform additive distribution. The values of probability $\alpha$ are $[0.001,0.01,0.1]$, and the uncertainty of the weights are $\delta=[25,50]$. We report the mean profit and the standard deviation of 30 independent runs for all algorithms. Each run is using $5*10^6$ fitness evaluations. A Kruskal-Wallis test \cite{corder2014nonparametric} with $95\%$ confidence interval integrated with the posterior Bonferroni test is used to compare multiple solutions.

In the following subsection, we consider all combinations of algorithms and operators. We set the $\beta$ in power-law distribution equal to $1.5$, which is the recommended value of $\beta$ from Doerr et al. \cite{Benjamin}. For each instance, we investigate different settings together with the difference between the uncertainty of weights and the chance-constrained probability. We report the results obtained by all algorithms with Chebyehsv inequality and Chernoff bound separately.

\subsection{Results for (1+1)~EA}

\begin{table}[t]
  \centering
  \caption{Statistic results of $(1+1)$~EA with Chernoff bound for instance eil101 with 500 items}
  \scalebox{1.0}{
  \makebox[\linewidth][c]{
  \tabcolsep=0.05cm
    \begin{tabular}{crrrrrrlrrl}
    \hline
          & \multicolumn{1}{l}{capacity} & \multicolumn{1}{l}{delta} & \multicolumn{1}{l}{alpha} &       & \multicolumn{3}{c}{Standard $(1+1)$~EA (1)} & \multicolumn{3}{c}{$(1+1)$~EA with HT (2)} \\
          &       &       &       &       & \multicolumn{1}{l}{Mean} & \multicolumn{1}{l}{Std} & stat  & \multicolumn{1}{l}{Mean} & \multicolumn{1}{l}{Std} & stat \\
          \hline
    \parbox[t]{2mm}{\multirow{12}{*}{\rotatebox[origin=c]{90}{bounded-strongly-correlated}}}  & 61447 & 25    & 0.001 &       & 77188.75 & 131.32 & 2(-)  & 77354.80 & 137.75 & $\mathbf{1(+)}$ \\
          &       &       & 0.01  &       & 77431.35 & 217.91 & 2(-)  & 77682.50 & 142.07 & $\mathbf{1(+)}$ \\
          &       &       & 0.1   &       & 77846.90 & 149.80 & 2(-)  & 78046.45 & 101.43 & $\mathbf{1(+)}$ \\
          &       & 50    & 0.001 &       & 75625.30 & 114.08 & 2(-)  & 75796.10 & 124.82 & $\mathbf{1(+)}$ \\
          &       &       & 0.01  &       & 76189.05 & 168.81 & 2(-)  & 76429.60 & 164.43 & $\mathbf{1(+)}$ \\
          &       &       & 0.1   &       & 76990.85 & 130.88 & 2(-)  & 77190.65 & 147.35 & $\mathbf{1(+)}$ \\
          & 162943 & 25    & 0.001 &       & 189768.50 & 176.14 & 2(-)  & 190192.30 & 109.27 & $\mathbf{1(+)}$ \\
          &       &       & 0.01  &       & 190136.65 & 146.38 & 2(-)  & 190435.60 & 152.04 & $\mathbf{1(+)}$ \\
          &       &       & 0.1   &       & 190668.95 & 164.61 & 2(-)  & 190889.80 & 138.44 & $\mathbf{1(+)}$ \\
          &       & 50    & 0.001 &       & 187930.55 & 200.91 & 2(-)  & 188244.65 & 87.04 & $\mathbf{1(+)}$ \\
          &       &       & 0.01  &       & 188636.25 & 185.64 & 2(-)  & 189002.70 & 157.02 & $\mathbf{1(+)}$ \\
          &       &       & 0.1   &       & 189560.60 & 185.47 & 2(-)  & 189882.45 & 134.64 & $\mathbf{1(+)}$ \\
          \hline
    \parbox[t]{2mm}{\multirow{12}{*}{\rotatebox[origin=c]{90}{uncorrelated}}} & 37686 & 25    & 0.001 &       & 85793.80 & 141.97 & 2(-)  & 85905.00 & 125.82 & $\mathbf{1(+)}$ \\
          &       &       & 0.01  &       & 86163.70 & 152.75 & 2(-)  & 86323.45 & 103.78 & $\mathbf{1(+)}$ \\
          &       &       & 0.1   &       & 86735.10 & 107.89 & 2(-)  & 86887.80 & 87.51 & $\mathbf{1(+)}$ \\
          &       & 50    & 0.001 &       & 83617.85 & 175.42 & 2(-)  & 83746.00 & 72.36 & $\mathbf{1(+)}$ \\
          &       &       & 0.01  &       & 84400.05 & 131.08 & 2(-)  & 84556.10 & 117.32 & $\mathbf{1(+)}$ \\
          &       &       & 0.1   &       & 85514.45 & 170.14 & 2(-)  & 85668.30 & 88.90 & $\mathbf{1(+)}$ \\
          & 93559 & 25    & 0.001 &       & 147538.15 & 105.60 & 2(-)  & 147693.65 & 45.21 & $\mathbf{1(+)}$ \\
          &       &       & 0.01  &       & 147931.80 & 164.88 & 2(-)  & 148048.80 & 64.67 & $\mathbf{1(+)}$ \\
          &       &       & 0.1   &       & 148371.20 & 101.71 & 2(-)  & 148515.65 & 76.13 & $\mathbf{1(+)}$ \\
          &       & 50    & 0.001 &       & 145675.90 & 73.28 & 2(-)  & 145767.40 & 88.57 & $\mathbf{1(+)}$ \\
          &       &       & 0.01  &       & 146381.05 & 123.87 & 2(-)  & 146478.65 & 65.55 & $\mathbf{1(+)}$ \\
          &       &       & 0.1   &       & 147311.05 & 98.65 & 2(-)  & 147450.10 & 78.27 & $\mathbf{1(+)}$ \\
          \hline
    \end{tabular}}}%
  \label{tab:onechernoff}%
\end{table}%

\begin{table}[t]
  \centering
  \caption{Statistic results of $(1+1)$~EA with Chebyshev's inequality for instance eil101 with 500 items}
  \scalebox{1.0}{
  \makebox[\linewidth][c]{
  \tabcolsep=0.05cm
    \begin{tabular}{crrrrrrlrrrl}
    \hline
          & \multicolumn{1}{l}{capacity} & \multicolumn{1}{l}{delta} & \multicolumn{1}{l}{alpha} &       & \multicolumn{3}{c}{ Standard $(1+1)$~EA (1)} &       & \multicolumn{3}{c}{$(1+1)$~EA with HT (2)} \\
          &       &       &       &       & \multicolumn{1}{l}{Mean} & \multicolumn{1}{l}{Std} & stat  &       & \multicolumn{1}{l}{Mean} & \multicolumn{1}{l}{Std} & stat \\
          \hline
    \parbox[t]{2mm}{\multirow{12}{*}{\rotatebox[origin=c]{90}{bounded-strongly-correlated}}}  & 61447 & 25    & 0.001 &       & 73845.45 & 154.22 & 2(-)  &       & 74030.20 & 106.77 & $\mathbf{1(+)}$ \\
          &       &       & 0.01  &       & 77118.95 & 177.20 & 2(-)  &       & 77288.75 & 112.60 & $\mathbf{1(+)}$ \\
          &       &       & 0.1   &       & 78184.80 & 166.21 & 2(-)  &       & 78353.85 & 143.99 & $\mathbf{1(+)}$ \\
          &       & 50    & 0.001 &       & 69136.30 & 210.29 & 2(-)  &       & 69394.60 & 121.15 & $\mathbf{1(+)}$ \\
          &       &       & 0.01  &       & 75619.85 & 159.61 & 2(-)  &       & 75829.30 & 129.03 & $\mathbf{1(+)}$ \\
          &       &       & 0.1   &       & 77712.80 & 201.23 & 2(-)  &       & 77987.45 & 124.92 & $\mathbf{1(+)}$ \\
          & 162943 & 25    & 0.001 &       & 185706.25 & 151.89 & 2(-)  &       & 185969.05 & 136.42 & $\mathbf{1(+)}$ \\
          &       &       & 0.01  &       & 189753.70 & 166.73 & 2(-)  &       & 190109.25 & 133.08 & $\mathbf{1(+)}$ \\
          &       &       & 0.1   &       & 191092.25 & 182.75 & 2(-)  &       & 191410.15 & 122.83 & $\mathbf{1(+)}$ \\
          &       & 50    & 0.001 &       & 179824.75 & 167.43 & 2(-)  &       & 180107.35 & 126.69 & $\mathbf{1(+)}$ \\
          &       &       & 0.01  &       & 187914.05 & 134.66 & 2(-)  &       & 188176.30 & 162.26 & $\mathbf{1(+)}$ \\
          &       &       & 0.1   &       & 190566.85 & 159.93 & 2(-)  &       & 190784.95 & 119.70 &
          $\mathbf{1(+)}$ \\
          \hline
     \parbox[t]{2mm}{\multirow{12}{*}{\rotatebox[origin=c]{90}{uncorrelated}}} & 37686 & 25    & 0.001 &       & 80931.05 & 206.08 & 2(-)  &       & 81114.65 & 108.74 & $\mathbf{1(+)}$ \\
          &       &       & 0.01  &       & 85710.55 & 181.84 & 2(-)  &       & 85857.05 & 130.49 & $\mathbf{1(+)}$ \\
          &       &       & 0.1   &       & 87306.40 & 174.52 & 2(-)  &       & 87405.90 & 134.22 & $\mathbf{1(+)}$ \\
          &       & 50    & 0.001 &       & 74345.30 & 232.35 & 2(-)  &       & 74483.10 & 152.61 & $\mathbf{1(+)}$ \\
          &       &       & 0.01  &       & 83497.20 & 132.89 & 2(-)  &       & 83685.30 & 117.64 & $\mathbf{1(+)}$ \\
          &       &       & 0.1   &       & 86617.35 & 146.49 & 2(-)  &       & 86761.15 & 83.79 & $\mathbf{1(+)}$ \\
          & 93559 & 25    & 0.001 &       & 143213.25 & 87.61 & 2(-)  &       & 143359.95 & 62.21 & $\mathbf{1(+)}$ \\
          &       &       & 0.01  &       & 147487.20 & 110.01 & 2(-)  &       & 147597.30 & 65.75 & $\mathbf{1(+)}$ \\
          &       &       & 0.1   &       & 148827.55 & 142.91 & 2(-)  &       & 148962.75 & 67.57 & $\mathbf{1(+)}$ \\
          &       & 50    & 0.001 &       & 137111.55 & 102.85 & 2(-)  &       & 137262.15 & 75.68 & $\mathbf{1(+)}$ \\
          &       &       & 0.01  &       & 145514.50 & 142.63 & 2(-)  &       & 145636.50 & 55.13 & $\mathbf{1(+)}$ \\
          &       &       & 0.1   &       & 148259.75 & 116.00 & 2(-)  &       & 148367.25 & 64.25 & $\mathbf{1(+)}$ \\
          \hline
    \end{tabular}}}%
  \label{tab:onechebyshev}%
\end{table}%

To show the differences between the evolutionary algorithms using the standard mutation operator and the heavy-tail mutation operator, we investigate the performance of $(1+1)$~EA using the heavy-tail mutation operator for solving the CCKP instances. Table~\ref{tab:onechernoff} and Table~\ref{tab:onechebyshev} list the average and standard deviation of profit for 30 independent runs concerning the probability estimate methods. For clarity, we use \textit{Standard $(1+1)$~EA} to represent the $(1+1)$~EA using standard mutation operator, and show the $(1+1)$~EA using heavy-tail mutation operator as \textit{$(1+1)$~EA with HT} in the tables. 

In Table~\ref{tab:onechernoff} and \ref{tab:onechebyshev}, the \textit{stat} column shows the rank of each algorithm in the instances. If two algorithms can be compared with each other significantly, X(+) denotes that the current algorithm is outperforming algorithm X. Besides, X(-) signifies that the current algorithm is worse than the algorithm X significantly. For example, the numbers 2(-) listed in the first row under the \textit{Standard $(1+1)$~EA(1)} mean that the current one is significantly worse than the solutions obtained by \textit{$(1+1)$~EA with HT mutation (2)}. 

The results in Table \ref{tab:onechernoff} and \ref{tab:onechebyshev} indicate that there is a significant difference between using the two mutation operators in the single-objective evolutionary algorithm. The $(1+1)$~EA with the heavy-tail mutation operator outperforms the standard $(1+1)$~EA in all cases. Moreover, in the most uncorrelated type of instances, the $(1+1)$~EA with heavy-tail mutations obtains solutions with a lower standard deviation in comparison with the other algorithm. In summary, the results show that the heavy-tail mutation operator leads to better performance when solving CCKP instances, which we have shown in bold at \textit{stat} columns.

\subsection{Results for $(\mu+1)$~EA}

The purpose of this section is to investigate the effectiveness of the problem-specific crossover operator associated with single-objective evolutionary algorithms. Here, we run the $(\mu+1)$~EA with the population size 10. To simplify the name of algorithms, we use the following notations: \textit{Standard $(\mu+1)$~EA} is the $(\mu+1)$~EA using standard mutation operator, \textit{$(\mu+1)$~EA with HT} is the $(\mu+1)$~EA using heavy-tail mutation operator and \textit{$(\mu+1)$~EA with HT and PS} represents the $(\mu+1)$~EA using the heavy-tail mutation operator and the problem-specific crossover operator.

\begin{table}[t]
 \centering
  \caption{Statistic results of $(\mu+1)$~EA with Chernoff bound for instance eil101 with 500 items}
  \scalebox{0.8}{
  \makebox[\linewidth][c]{
  \tabcolsep=0.05cm
    \begin{tabular}{crrrrrrlrrrlrrrl}
    \hline
          & \multicolumn{1}{l}{capacity} & \multicolumn{1}{l}{delta} & \multicolumn{1}{l}{alpha} &       & \multicolumn{3}{c}{Standard $(\mu+1)$~EA (3)} &       & \multicolumn{3}{c}{$(\mu+1)$~EA with HT (4)} &       & \multicolumn{3}{c}{$(\mu+1)$~EA with HT and PS (5)} \\
          &       &       &       &       & \multicolumn{1}{l}{Mean} & \multicolumn{1}{l}{Std} & stat  &       & \multicolumn{1}{l}{Mean} & \multicolumn{1}{l}{Std} & stat  &       & \multicolumn{1}{l}{Mean} & \multicolumn{1}{l}{Std} & stat \\
          \hline
     \parbox[t]{2mm}{\multirow{12}{*}{\rotatebox[origin=c]{90}{bounded-strongly-correlated}}}  & 61447 & 25    & 0.001 &       & 77112.27 & 182.05 & 4(-),5(-) &       & 77350.17 & 108.33 & 3(+),5(-) &       & 77518.20 & 104.44 & $\mathbf{3(+),4(+)}$ \\
          &       &       & 0.01  &       & 77413.97 & 188.46 & 4(-),5(-) &       & 77646.67 & 119.40 & 3(+),5(-) &       & 77811.80 & 107.17 & $\mathbf{3(+),4(+)}$ \\
          &       &       & 0.1   &       & 77787.77 & 138.06 & 4(-),5(-) &       & 78000.57 & 133.04 & 3(+),5(-) &       & 78197.40 & 142.42 & $\mathbf{3(+),4(+)}$ \\
          &       & 50    & 0.001 &       & 75562.23 & 201.47 & 4(-),5(-) &       & 75816.83 & 141.98 & 3(+),5(-) &       & 75934.03 & 102.39 & $\mathbf{3(+),4(+)}$ \\
          &       &       & 0.01  &       & 76178.43 & 156.15 & 4(-),5(-) &       & 76455.20 & 113.76 & 3(+),5(-) &       & 76559.90 & 138.41 & $\mathbf{3(+),4(+)}$ \\
          &       &       & 0.1   &       & 76948.67 & 177.82 & 4(-),5(-) &       & 77193.60 & 136.70 & 3(+),5(-) &       & 77296.53 & 148.22 & $\mathbf{3(+),4(+)}$ \\
          & 162943 & 25    & 0.001 &       & 189795.90 & 164.86 & 4(-),5(-) &       & 190130.60 & 142.84 & 3(+),5(-) &       & 190418.70 & 155.28 & $\mathbf{3(+),4(+)}$ \\
          &       &       & 0.01  &       & 190179.80 & 150.85 & 4(-),5(-) &       & 190523.13 & 118.16 & 3(+),5(-) &       & 190814.40 & 130.31 & $\mathbf{3(+),4(+)}$ \\
          &       &       & 0.1   &       & 190569.93 & 194.37 & 4(-),5(-) &       & 190917.27 & 122.95 & 3(+),5(-) &       & 191286.05 & 127.97 & $\mathbf{3(+),4(+)}$ \\
          &       & 50    & 0.001 &       & 188027.17 & 138.12 & 4(-),5(-) &       & 188308.87 & 129.41 & 3(+),5(-) &       & 188575.15 & 146.49 & $\mathbf{3(+),4(+)}$ \\
          &       &       & 0.01  &       & 188690.86 & 139.27 & 4(-),5(-) &       & 189016.46 & 112.01 & 3(+),5(-) &       & 189287.95 & 108.62 & $\mathbf{3(+),4(+)}$ \\
          &       &       & 0.1   &       & 189574.83 & 184.21 & 4(-),5(-) &       & 189862.87 & 118.61 & 3(+),5(-) &       & 190207.65 & 130.52 & $\mathbf{3(+),4(+)}$ \\
          \hline
     \parbox[t]{2mm}{\multirow{12}{*}{\rotatebox[origin=c]{90}{uncorrelated}}} & 37686 & 25    & 0.001 &       & 85788.63 & 108.84 & 4(-),5(-) &       & 85938.83 & 113.10 & 3(+),5(-) &       & 85968.47 & 82.01 & $\mathbf{3(+),4(+)}$ \\
          &       &       & 0.01  &       & 86198.93 & 147.91 & 4(-),5(-) &       & 86331.03 & 109.54 & 3(+),5(-) &       & 86395.43 & 87.11 & $\mathbf{3(+),4(+)}$ \\
          &       &       & 0.1   &       & 86784.87 & 133.35 & 4(-),5(-) &       & 86887.37 & 104.28 & 3(+),5(-) &       & 86925.50 & 88.78 & $\mathbf{3(+),4(+)}$ \\
          &       & 50    & 0.001 &       & 83586.80 & 204.58 & 4(-),5(-) &       & 83745.63 & 101.20 & 3(+),5(-) &       & 83819.10 & 91.67 & $\mathbf{3(+),4(+)}$ \\
          &       &       & 0.01  &       & 84476.20 & 143.76 & 4(-),5(-) &       & 84572.10 & 104.82 & 3(+),5(-) &       & 84641.60 & 126.83 & $\mathbf{3(+),4(+)}$ \\
          &       &       & 0.1   &       & 85521.30 & 166.74 & 4(-),5(-) &       & 85654.07 & 105.95 & 3(+),5(-) &       & 85717.10 & 97.85 & $\mathbf{3(+),4(+)}$ \\
          & 93559 & 25    & 0.001 &       & 147537.37 & 128.76 & 4(-),5(-) &       & 147683.03 & 70.76 & 3(+),5(-) &       & 147745.95 & 76.22 & $\mathbf{3(+),4(+)}$ \\
          &       &       & 0.01  &       & 147923.97 & 98.21 & 4(-),5(-) &       & 148032.67 & 65.28 & 3(+),5(-) &       & 148126.05 & 61.35 & $\mathbf{3(+),4(+)}$ \\
          &       &       & 0.1   &       & 148388.83 & 106.95 & 4(-),5(-) &       & 148513.17 & 74.87 & 3(+),5(-) &       & 148604.60 & 53.52 & $\mathbf{3(+),4(+)}$ \\
          &       & 50    & 0.001 &       & 145648.43 & 90.76 & 4(-),5(-) &       & 145770.03 & 65.78 & 3(+),5(-) &       & 145838.10 & 59.74 & $\mathbf{3(+),4(+)}$ \\
          &       &       & 0.01  &       & 146379.80 & 86.76 & 4(-),5(-) &       & 146513.47 & 57.61 & 3(+),5(-) &       & 146572.60 & 47.19 & $\mathbf{3(+),4(+)}$ \\
          &       &       & 0.1   &       & 147316.30 & 91.46 & 4(-),5(-) &       & 147414.76 & 85.65 & 3(+),5(-) &       & 147497.10 & 60.68 & $\mathbf{3(+),4(+)}$ \\
          \hline
    \end{tabular}}}%
  \label{tab:muchernoff}%
\end{table}%

\begin{table}[t]
  \centering
  \caption{Statistic results of $(\mu+1)$~EA with Chebyshev's inequality for instance eil101 with 500 items}
  \scalebox{0.8}{
  \makebox[\linewidth][c]{
  \tabcolsep=0.05cm
    \begin{tabular}{crrrrrrlrrrlrrrl}
    \hline
          & \multicolumn{1}{l}{capacity} & \multicolumn{1}{l}{delta} & \multicolumn{1}{l}{alpha} &       & \multicolumn{3}{c}{ Standard $(\mu+1)$~EA (3)} &       & \multicolumn{3}{c}{$(\mu+1)$~EA with HT (4)} &       & \multicolumn{3}{c}{$(\mu+1)$~EA with HT and PS (5)} \\
          &       &       &       &       & \multicolumn{1}{l}{Mean} & \multicolumn{1}{l}{Std} & stat  &       & \multicolumn{1}{l}{Mean} & \multicolumn{1}{l}{Std} & stat  &       & \multicolumn{1}{l}{Mean} & \multicolumn{1}{l}{Std} & stat \\
          \hline
     \parbox[t]{2mm}{\multirow{12}{*}{\rotatebox[origin=c]{90}{bounded-strongly-correlated}}} & 61447 & 25    & 0.001 &       & 73881.73 & 175.53 & 4(-),5(-) &       & 74005.90 & 137.56 & 3(+),5(-) &       & 74076.60 & 153.03 & $\mathbf{3(+),4(+)}$ \\
          &       &       & 0.01  &       & 77141.50 & 161.02 & 4(-),5(-) &       & 77376.87 & 105.29 & 3(+),5(-) &       & 77515.70 & 106.62 & $\mathbf{3(+),4(+)}$ \\
          &       &       & 0.1   &       & 78171.23 & 212.15 & 4(-),5(-) &       & 78000.57 & 166.67 & 3(+),5(-) &       & 78620.00 & 135.42 & $\mathbf{3(+),4(+)}$ \\
          &       & 50    & 0.001 &       & 69178.33 & 151.69 & 4(-),5(-) &       & 69393.30 & 103.48 & 3(+),5(-) &       & 69439.10 & 110.33 & $\mathbf{3(+),4(+)}$ \\
          &       &       & 0.01  &       & 75661.87 & 138.77 & 4(-),5(-) &       & 76455.20 & 121.03 & 3(+),5(-) &       & 75968.90 & 100.49 & $\mathbf{3(+),4(+)}$ \\
          &       &       & 0.1   &       & 77735.60 & 201.61 & 4(-),5(-) &       & 77982.27 & 136.70 & 3(+),5(-) &       & 78157.37 & 98.79 & $\mathbf{3(+),4(+)}$ \\
          & 162943 & 25    & 0.001 &       & 185726.67 & 121.77 & 4(-),5(-) &       & 186003.77 & 117.99 & 3(+),5(-) &       & 186232.25 & 111.68 & $\mathbf{3(+),4(+)}$ \\
          &       &       & 0.01  &       & 189717.60 & 163.28 & 4(-),5(-) &       & 190063.07 & 152.09 & 3(+),5(-) &       & 190416.70 & 117.58 & $\mathbf{3(+),4(+)}$ \\
          &       &       & 0.1   &       & 191088.17 & 145.18 & 4(-),5(-) &       & 191362.76 & 117.11 & 3(+),5(-) &       & 191808.15 & 109.23 & $\mathbf{3(+),4(+)}$ \\
          &       & 50    & 0.001 &       & 179826.70 & 163.99 & 4(-),5(-) &       & 180083.90 & 118.26 & 3(+),5(-) &       & 180218.70 & 106.71 & $\mathbf{3(+),4(+)}$ \\
          &       &       & 0.01  &       & 187871.70 & 153.35 & 4(-),5(-) &       & 188232.13 & 120.62 & 3(+),5(-) &       & 188546.80 & 139.90 & $\mathbf{3(+),4(+)}$ \\
          &       &       & 0.1   &       & 190485.00 & 166.26 & 4(-),5(-) &       & 190831.97 & 131.35 & 3(+),5(-) &       & 191183.70 & 160.72 & $\mathbf{3(+),4(+)}$ \\
          \hline
     \parbox[t]{2mm}{\multirow{12}{*}{\rotatebox[origin=c]{90}{uncorrelated}}}  & 37686 & 25    & 0.001 &       & 81049.37 & 140.15 & 4(-),5(-) &       & 81109.53 & 131.12 & 3(+),5(-) &       & 81184.30 & 109.41 & $\mathbf{3(+),4(+)}$ \\
          &       &       & 0.01  &       & 85798.93 & 121.88 & 4(-),5(-) &       & 85898.80 & 97.51 & 3(+),5(-) &       & 85959.10 & 84.36 & $\mathbf{3(+),4(+)}$ \\
          &       &       & 0.1   &       & 87322.67 & 142.69 & 4(-),5(-) &       & 87449.93 & 97.28 & 3(+),5(-) &       & 87486.60 & 82.56 & $\mathbf{3(+),4(+)}$ \\
          &       & 50    & 0.001 &       & 74378.33 & 174.08 & 4(-),5(-) &       & 74498.87 & 119.64 & 3(+),5(-) &       & 74597.45 & 120.22 & $\mathbf{3(+),4(+)}$ \\
          &       &       & 0.01  &       & 83554.13 & 153.41 & 4(-),5(-) &       & 83665.77 & 104.76 & 3(+),5(-) &       & 83723.30 & 97.05 & $\mathbf{3(+),4(+)}$ \\
          &       &       & 0.1   &       & 86634.27 & 150.50 & 4(-),5(-) &       & 86801.60 & 78.42 & 3(+),5(-) &       & 86872.40 & 88.62 & $\mathbf{3(+),4(+)}$ \\
          & 93559 & 25    & 0.001 &       & 143256.20 & 97.24 & 4(-),5(-) &       & 143350.47 & 72.07 & 3(+),5(-) &       & 143420.45 & 58.82 & $\mathbf{3(+),4(+)}$ \\
          &       &       & 0.01  &       & 147521.47 & 81.86 & 4(-),5(-) &       & 148032.67 & 75.09 & 3(+),5(-) &       & 147703.15 & 53.83 & $\mathbf{3(+),4(+)}$ \\
          &       &       & 0.1   &       & 148900.87 & 88.43 & 4(-),5(-) &       & 149009.90 & 81.65 & 3(+),5(-) &       & 149083.10 & 51.14 & $\mathbf{3(+),4(+)}$ \\
          &       & 50    & 0.001 &       & 137174.83 & 102.32 & 4(-),5(-) &       & 137303.53 & 64.42 & 3(+),5(-) &       & 137317.90 & 53.63 & $\mathbf{3(+),4(+)}$ \\
          &       &       & 0.01  &       & 145525.53 & 111.41 & 4(-),5(-) &       & 145645.90 & 81.14 & 3(+),5(-) &       & 145719.65 & 37.79 & $\mathbf{3(+),4(+)}$ \\
          &       &       & 0.1   &       & 148298.47 & 119.97 & 4(-),5(-) &       & 148409.80 & 62.97 & 3(+),5(-) &       & 148491.60 & 40.45 & $\mathbf{3(+),4(+)}$ \\
          \hline
    \end{tabular}}}%
  \label{tab:muchebyshev}%
\end{table}%

Table \ref{tab:muchernoff} and \ref{tab:muchebyshev} list the results when using Chernoff bound and Chebyshev's inequality to estimate the constraint violation probability of a CCKP solution separately. As can be seen from the tables, the performance of using the heavy-tail mutation operator is significantly better than using the standard mutation operator on all instances. Therefore the conclusion is the same as for the $(1+1)$~EA. Another insight of these tables can be drawn from the values of the columns \textit{$(\mu+1)$~EA with HT (4)} and \textit{$(\mu+1)$~EA with HT and PS (5)}. We can clearly see that the results obtained by \textit{$(\mu+1)$~EA with HT and PS (5)} are significantly better than \textit{$(\mu+1)$~EA with HT (4)}. It shows the effectiveness of the PS crossover operator when solving the CCKP instances in single-objective evolutionary algorithms compared to mutation only. 

Moreover, by comparing the values of the corresponding columns in the Table \ref{tab:onechernoff} and \ref{tab:muchernoff}, and in Table \ref{tab:onechebyshev} and \ref{tab:muchebyshev} respectively according to the probability tails. It can be seen that in both estimated methods, the results in \textit{$(\mu+1)$~EA with HT and PS (5)} are better than the other combinations of algorithms and operators. We bold the statistic results in tables. In summary, the results in this section indicate that the heavy-tail mutation operator and the problem-specific crossover operator are active in single-objective evolutionary algorithms when dealing with the CCKP instances. The next section, therefore, moves on to discuss the performance of these operators in multi-objective approaches.

\section{Multi-Objective Approaches}
\label{multiapp}

In this section, we introduce a new multi-objective model for the chance-constrained knapsack problem. The model considers both feasible solutions and the second type of infeasible solutions than mentioned in Section \ref{surrogate}. Then we apply the new model to GSEMO previous considered for CCKP~\cite{Xie}. GSEMO can generate a Pareto front with both feasible solutions and infeasible solutions. For further investigation of our multi-objective optimization, we also apply the Non-dominated Sorting Genetic Algorithm (NSGA-II) \cite{996017}, which is a state of the art multi-objective EA when dealing with two objectives. We run NSGA-II using 20 as a population size with Chebyshev's inequality and Chernoff bound, respectively. 

\subsection{New Multi-Objective Model for CCKP}
\label{multimodel}

To keep more diversity in the solution space, the new model makes other solutions dominate the infeasible solutions that the expected weight of selection items is overloading the capacity. The difference between the new multi-objective model and the old multi-objective model in \cite{Xie} is that the old model made all feasible solutions dominate all infeasible solutions. The fitness functions of this new model are proposed as follow. 

\begin{equation}
g_1(X)=\left\{
\begin{array}{rcl}
Prob(W\geq C) & & {E_W (X)<C}\\
1+(E_W (X)-C)& & {E_W (X)\geq C}
\label{g2x}
\end{array} \right.
\end{equation}

\begin{equation}
g_2(X)=\left\{
\begin{array}{lcl}
\sum_{i=1}^n {p_i x_i} & & { g_1 (X)\leq 1}\\
-1 & & {g_1 (X) >1}
\label{g1x}
\end{array} \right.
\end{equation}

The first function calculates the probability of a solution by overloading the capacity of the knapsack, and it forces the probability of an infeasible solution whose expected weight exceeds the capacity larger than $1$. The second fitness function is the objective of the chance-constrained knapsack problem. It calculates the profit of feasible solutions that the probability less than $\alpha$ and infeasible solutions with probability more than $\alpha$ but less than in $1$. We say solution $Y$ dominates solution $X$ w.r.t. ${g}$, denoted by $Y \succcurlyeq X$, iff $g_1(Y)\leq g_1(X) \land g_2(Y) \geq g_2 (X)$. 

The objective function ${g_2}$ guarantees that the search process is guided towards all considering solutions, that trade-offs in terms of confidence level and profit are computed for the solutions in the Pareto front. However, even the algorithm can store feasible solutions and a bunch of infeasible solutions, we output the best feasible solution in every iteration. 

\subsection{Evolutionary Algorithms}
The first multi-objective approach we consider here is a simple multi-objective evolutionary algorithm as a baseline multi-objective approach, which is named for Global Simple Evolutionary Multi-Objective Optimizer (GSEMO) \cite{Giel}. The GSEMO, given in Algorithm \ref{alg:multiobj} works like $(1+1)$~EA, starts with a random search point $x\in\{0,1\}^n$. In the mutation step, flip each bit with a probability $1/n$, but the algorithm stores a set of solutions in the main optimization loop where any solution does not dominate each other. 

\begin{algorithm}[tp]
\caption{GSEMO}
\begin{algorithmic}[1]
\STATE Choose $x \in \{0,1\}^n$ uniformly at random \;
\STATE $S\leftarrow \{x\}$;
\WHILE{stopping criterion not met}
\STATE choose $x\in S$ uniformly at random;
\STATE $y\leftarrow$ flip each bit of $x$ independently with probability of $\frac{1}{n}$;
\IF{($\not\exists w \in S: w \succcurlyeq_{GSEMO} y$)}
\STATE $S \leftarrow (S \cup \{y\})\backslash \{z\in S \vert y \succcurlyeq_{GSEMO} z\}$;
\ENDIF
\ENDWHILE
\end{algorithmic}
\label{alg:multiobj}
\end{algorithm}

We also apply the Non-dominated Sorting Genetic Algorithm (NSGA-II), which is a state-of-art multi-objective EA. NSGA-II was proposed by Deb et al. \cite{996017}. For the detail of this algorithm, we refer to read the paper \cite{996017}. We run NSGA-II with a population size of 20 using Chebyshev and Chernoff inequality tails respectively, and compare the performance of GSEMO and NSGA-II at the end of this section.
\begin{table}[t]
  \centering
  \caption{Statistic results of GSEMO with Chernoff bound for the instance eil101 with 500 items}
  \scalebox{0.8}{
  \makebox[\linewidth][c]{
  \tabcolsep=0.05cm
    \begin{tabular}{crrrrrrlrrrlrrrl}
    \hline
          & \multicolumn{1}{l}{capacity} & \multicolumn{1}{l}{delta} & \multicolumn{1}{l}{alpha} &       & \multicolumn{3}{c}{GSEMO with old model (6)} &       & \multicolumn{3}{c}{GSEMO with new model (7)} &       & \multicolumn{3}{c}{GSEMO with new model and PS (8)} \\
          &       &       &       &       & \multicolumn{1}{l}{Mean} & \multicolumn{1}{l}{Std} & stat  &       & \multicolumn{1}{l}{Mean} & \multicolumn{1}{l}{Std} & stat  &       & \multicolumn{1}{l}{Mean} & \multicolumn{1}{l}{Std} & stat \\
  \hline
       \parbox[t]{2mm}{\multirow{12}{*}{\rotatebox[origin=c]{90}{bounded-strongly-correlated}}} & 61447 & 25    & 0.001 &       & 77907.23 & 28.13 & 7(-),8(-) &       & 77921.57 & 30.63 & 6(+),8(-) &       & 77955.00 & 0.00  & 6(+),7(+) \\
          &       &       & 0.01  &       & 78242.20 & 1.86  &       &       & 78242.47 & 0.81  &       &       & 78243.00 & 0.00  &  \\
          &       &       & 0.1   &       & 78592.00 & 15.55 & 8(-)  &       & 78596.80 & 20.83 & 8(-)  &       & 78649.00 & 0.00  & 6(+),7(+) \\
          &       & 50    & 0.001 &       & 76341.33 & 6.83  &       &       & 76344.73 & 0.69  &       &       & 76345.00 & 0.00  &  \\
          &       &       & 0.01  &       & 76955.33 & 9.48  & 7(-),8(-) &       & 76960.47 & 1.17  & 6(+)  &       & 76961.00 & 0.00  & 6(+) \\
          &       &       & 0.1   &       & 77721.87 & 2.45  &       &       & 77722.80 & 0.81  &       &       & 77723.00 & 0.00  &  \\
          & 162943 & 25    & 0.001 &       & 190899.33 & 16.90 & 8(-)  &       & 190901.23 & 8.68  & 8(-)  &       & 190909.00 & 0.79  & 6(+),7(+) \\
          &       &       & 0.01  &       & 191244.80 & 3.25  &       &       & 191246.00 & 2.00  &       &       & 191245.60 & 0.55  &  \\
          &       &       & 0.1   &       & 191682.97 & 24.79 & 8(-)  &       & 191674.40 & 1.14  & 8(-)  &       & 191759.00 & 0.00  & 6(+),7(+) \\
          &       & 50    & 0.001 &       & 188977.60 & 25.98 & 7(-),8(-) &       & 188985.40 & 9.24  & 6(+),8(-) &       & 188996.29 & 2.45  & 6(+),7(+) \\
          &       &       & 0.01  &       & 189687.60 & 25.64 & 8(-)  &       & 189687.80 & 24.96 & 8(-)  &       & 189748.27 & 7.43  & 6(+),7(+) \\
          &       &       & 0.1   &       & 190604.00 & 3.39  & 7(-),8(-) &       & 190609.80 & 8.64  & 6(+),8(-) &       & 190686.27 & 5.35  & 6(+),7(+) \\
          \hline
       \parbox[t]{2mm}{\multirow{12}{*}{\rotatebox[origin=c]{90}{uncorrelated}}} & 37686 & 25    & 0.001 &       & 86246.17 & 6.98  &       &       & 86247.00 & 5.00  &       &       & 86248.00 & 0.00  &  \\
          &       &       & 0.01  &       & 86653.47 & 1.66  &       &       & 86654.00 & 0.00  &       &       & 86654.00 & 0.00  &  \\
          &       &       & 0.1   &       & 87213.89 & 2.36  &       &       & 87214.00 & 0.00  &       &       & 87214.00 & 0.00  &  \\
          &       & 50    & 0.001 &       & 84065.07 & 8.17  &       &       & 84069.70 & 0.47  &       &       & 84070.00 & 0.00  &  \\
          &       &       & 0.01  &       & 84909.97 & 10.33 & 8(-)  &       & 84914.07 & 7.65  &       &       & 84917.00 & 0.00  & 6(+) \\
          &       &       & 0.1   &       & 85965.83 & 6.30  &       &       & 85966.47 & 2.37  &       &       & 85967.00 & 0.00  &  \\
          & 93559 & 25    & 0.001 &       & 147885.00 & 3.36  &       &       & 147888.60 & 0.89  &       &       & 147889.00 & 0.00  &  \\
          &       &       & 0.01  &       & 148252.40 & 4.75  &       &       & 148256.20 & 0.84  &       &       & 148257.00 & 0.00  &  \\
          &       &       & 0.1   &       & 148720.55 & 3.83  & 8(-)  &       & 148724.60 & 2.51  &       &       & 148726.60 & 0.89  & 6(+) \\
          &       & 50    & 0.001 &       & 145964.00 & 5.00  & 8(-)  &       & 145965.60 & 1.34  & 8(-)  &       & 145973.58 & 6.89  & 6(+),7(+) \\
          &       &       & 0.01  &       & 146696.00 & 8.80  & 8(-)  &       & 146694.00 & 10.12 & 8(-)  &       & 146710.00 & 0.00  & 6(+),7(+) \\
          &       &       & 0.1   &       & 147623.80 & 10.06 & 7(-),8(-) &       & 147632.20 & 7.60  & 6(+),8(-) &       & 147642.00 & 0.00  & 6(+),7(+) \\
          \hline
    \end{tabular}}}%
  \label{tab:gsemoChernoff}%
\end{table}%

\begin{table}[htbp]
  \centering
  \caption{Statistic results of GSEMO with Chebyshev's inequality for the instance eil101 with 500 items}
  \scalebox{0.8}{
  \makebox[\linewidth][c]{
  \tabcolsep=0.05cm
    \begin{tabular}{crrrrrrlrrrlrrrl}
    \hline
          & \multicolumn{1}{l}{capacity} & \multicolumn{1}{l}{delta} & \multicolumn{1}{l}{alpha} &       & \multicolumn{3}{c}{GSEMO with old model (6)} &       & \multicolumn{3}{c}{GSEMO with new model (7)} &       & \multicolumn{3}{c}{GSEMO with new model and PS (8)} \\
          &       &       &       &       & \multicolumn{1}{l}{Mean} & \multicolumn{1}{l}{Std} & stat  &       & \multicolumn{1}{l}{Mean} & \multicolumn{1}{l}{Std} & stat  &       & \multicolumn{1}{l}{Mean} & \multicolumn{1}{l}{Std} & stat \\
          \hline
   \parbox[t]{2mm}{\multirow{12}{*}{\rotatebox[origin=c]{90}{bounded-strongly-correlated}}} & 61447 & 25    & 0.001 &       & 74505.00 & 11.76 & 7(-),8(-) &       & 74514.80 & 0.81  & 6(+)  &       & 74515.00 & 0.00  & 6(+) \\
          &       &       & 0.01  &       & 77882.23 & 5.67  & 8(-)  &       & 77885.23 & 5.41  & 8(-)  &       & 77953.00 & 0.00  & 6(+),7(+) \\
          &       &       & 0.1   &       & 79026.23 & 3.73  &       &       & 79028.10 & 1.06  &       &       & 79029.00 & 0.00  &  \\
          &       & 50    & 0.001 &       & 69852.37 & 29.36 & 7(-),8(-) &       & 69900.60 & 8.69  & 6(+),8(-) &       & 69925.00 & 0.00  & 6(+),7(+) \\
          &       &       & 0.01  &       & 76327.47 & 13.59 & 7(-),8(-) &       & 76336.13 & 2.01  & 6(+)  &       & 76337.00 & 0.00  & 6(+) \\
          &       &       & 0.1   &       & 78517.57 & 3.83  & 8(-)  &       & 78519.60 & 4.61  & 8(-)  &       & 78525.00 & 0.00  & 6(+),7(+) \\
          & 162943 & 25    & 0.001 &       & 186609.20 & 21.00 & 7(-),8(-) &       & 186651.20 & 18.32 & 6(+),8(-) &       & 186666.00 & 5.10  & 6(+),7(+) \\
          &       &       & 0.01  &       & 190795.60 & 27.49 & 7(-),8(-) &       & 190800.80 & 32.41 & 6(+),8(-) &       & 190866.00 & 8.78  & 6(+),7(+) \\
          &       &       & 0.1   &       & 192177.97 & 35.88 & 7(-),8(-) &       & 192202.00 & 3.74  & 6(+),8(-) &       & 192207.20 & 0.45  & 6(+),7(+) \\
          &       & 50    & 0.001 &       & 180661.00 & 36.85 & 7(-),8(-) &       & 180696.80 & 10.55 & 6(+),8(-) &       & 180765.80 & 8.23  & 6(+),7(+) \\
          &       &       & 0.01  &       & 188857.40 & 39.80 & 7(-),8(-) &       & 188888.40 & 27.14 & 6(+),8(-) &       & 188902.00 & 8.17  & 6(+),7(+) \\
          &       &       & 0.1   &       & 191575.00 & 4.58  & 8(-)  &       & 191576.00 & 1.87  & 8(-)  &       & 191581.33 & 1.65  & 6(+),7(+) \\
          \hline
    \parbox[t]{2mm}{\multirow{12}{*}{\rotatebox[origin=c]{90}{uncorrelated}}}   & 37686 & 25    & 0.001 &       & 81479.37 & 3.39  &       &       & 81480.73 & 1.38  &       &       & 81480.20 & 1.79  &  \\
          &       &       & 0.01  &       & 86206.30 & 1.66  &       &       & 86208.77 & 4.48  &       &       & 86210.00 & 0.00  &  \\
          &       &       & 0.1   &       & 87728.77 & 4.36  &       &       & 87730.17 & 2.09  &       &       & 87732.00 & 0.00  &  \\
          &       & 50    & 0.001 &       & 74869.00 & 3.81  &       &       & 74868.80 & 4.63  &       &       & 74870.00 & 0.00  &  \\
          &       &       & 0.01  &       & 84013.50 & 6.40  & 7(-),8(-) &       & 84019.90 & 4.30  & 6(+)  &       & 84022.00 & 0.00  & 6(+) \\
          &       &       & 0.1   &       & 87066.73 & 5.30  &       &       & 87068.70 & 0.47  &       &       & 87069.00 & 0.00  &  \\
          & 93559 & 25    & 0.001 &       & 143559.20 & 7.91  & 8(-)  &       & 143557.60 & 6.58  & 8(-)  &       & 143568.00 & 13.11 & 6(+),7(+) \\
          &       &       & 0.01  &       & 147816.35 & 7.26  & 8(-)  &       & 147821.20 & 2.77  &       &       & 147825.40 & 1.34  & 6(+) \\
          &       &       & 0.1   &       & 149211.15 & 5.85  & 8(-)  &       & 149211.80 & 5.85  & 8(-)  &       & 149218.00 & 0.00  & 6(+),7(+) \\
          &       & 50    & 0.001 &       & 137517.60 & 0.89  &  &       & 137515.60 & 2.19  &   &       & 137517.78 & 8.05  &  \\
          &       &       & 0.01  &       & 145843.40 & 2.97  & 7(-),8(-) &       & 145849.60 & 3.78  & 6(+),8(-) &       & 145855.14 & 22.98 & 6(+),7(+) \\
          &       &       & 0.1   &       & 148603.60 & 3.29  & 7(-),8(-) &       & 148612.80 & 3.42  & 6(+)  &       & 148614.90 & 2.78  & 6(+) \\
          \hline
    \end{tabular}}}%
  \label{tab:gsemoChebyshev}%
\end{table}%

\subsection{Experimental Results for GSEMO}

\begin{sidewaystable*}[h]
  \centering
  \caption{Statistic results of NSGA-II with Chernoff bound for instance eil101 with 500 items}
  \scalebox{0.8}{
  \makebox[\linewidth][c]{
  \tabcolsep=0.05cm
    \begin{tabular}{crrrrrrlrrrlrrrlrrrl}
    \hline
          & \multicolumn{1}{l}{capacity} & \multicolumn{1}{l}{delta} & \multicolumn{1}{l}{alpha} &       & \multicolumn{3}{c}{NSGA-II with old and uniform (9)} &       & \multicolumn{3}{c}{NSGA-II with old and PS (10)} &       & \multicolumn{3}{c}{NSGA-II with new and uniform (11)} &       & \multicolumn{3}{c}{NSGA-II with new and PS (12)} \\
          &       &       &       &       & \multicolumn{1}{l}{Mean} & \multicolumn{1}{l}{Std} & stat  &       & \multicolumn{1}{l}{Mean} & \multicolumn{1}{l}{Std} & stat  &       & \multicolumn{1}{l}{Mean} & \multicolumn{1}{l}{Std} & stat  &       & \multicolumn{1}{l}{Mean} & \multicolumn{1}{l}{Std} & stat \\
          \hline
    \parbox[t]{2mm}{\multirow{12}{*}{\rotatebox[origin=c]{90}{bounded-strongly-correlated}}} & 61447 & 25    & 0.001 &       & 77504.93 & 148.92 & 10(-),11(-),12(-) &       & 77836.33 & 59.83 & 9(+),11(+),12(-) &       & 77724.13 & 97.36 & 9(+),10(-),12(-) &       & 77914.33 & 40.90 & 9(+),10(+),11(+) \\
          &       &       & 0.01  &       & 77878.60 & 145.69 & 10(-),11(-),12(-) &       & 78154.63 & 60.99 & 9(+),11(+),12(-) &       & 78078.10 & 61.79 & 9(+),10(-),12(-) &       & 78220.37 & 30.01 & 9(+),10(+),11(+) \\
          &       &       & 0.1   &       & 78267.03 & 120.50 & 10(-),11(-),12(-) &       & 78558.20 & 49.06 & 9(+),11(+),12(-) &       & 78464.67 & 85.56 & 9(+),10(-),12(-) &       & 78595.33 & 30.28 & 9(+),10(+),11(+) \\
          &       & 50    & 0.001 &       & 75911.17 & 145.99 & 10(-),11(-),12(-) &       & 76252.93 & 67.80 & 9(+),11(-),12(-) &       & 76698.10 & 89.40 & 9(+),10(+),12(-) &       & 76932.93 & 23.63 & 9(+),10(+),11(+) \\
          &       &       & 0.01  &       & 76581.50 & 126.37 & 10(-),11(-),12(-) &       & 76883.90 & 61.50 & 9(+),11(+),12(-) &       & 76698.10 & 89.40 & 9(+),10(-),12(-) &       & 76932.93 & 23.63 & 9(+),10(+),11(+) \\
          &       &       & 0.1   &       & 77439.80 & 101.89 & 10(-),11(-),12(-) &       & 77699.27 & 29.82 & 9(+),11(+),12(-) &       & 77534.43 & 72.53 & 9(+),10(-),12(-) &       & 77712.27 & 8.31  & 9(+),10(+),11(+) \\
          & 162943 & 25    & 0.001 &       & 190285.93 & 190.97 & 10(-),11(-),12(-) &       & 190841.83 & 54.12 & 9(+),11(+),12(-) &       & 190532.70 & 119.81 & 9(+),10(-),12(-) &       & 190888.73 & 12.46 & 9(+),10(+),11(+) \\
          &       &       & 0.01  &       & 190686.43 & 182.27 & 10(-),11(-),12(-) &       & 191209.57 & 42.33 & 9(+),11(+),12(-) &       & 190937.13 & 88.80 & 9(+),10(-),12(-) &       & 191227.20 & 19.37 & 9(+),10(+),11(+) \\
          &       &       & 0.1   &       & 191149.50 & 138.73 & 10(-),11(-),12(-) &       & 191549.50 & 18.71 & 9(+),11(+),12(-) &       & 191398.70 & 81.26 & 9(+),10(-),12(-) &       & 191693.53 & 42.38 & 9(+),10(+),11(+) \\
          &       & 50    & 0.001 &       & 188328.37 & 193.81 & 10(-),11(+),12(-) &       & 188952.87 & 41.35 & 9(+),11(+) &       & 188260.07 & 146.80 & 9(-),10(-),12(-) &       & 188953.47 & 12.58 & 9(+),10(+) \\
          &       &       & 0.01  &       & 189111.80 & 175.58 & 10(-),11(+),12(-) &       & 189686.80 & 49.97 & 9(+),11(+),12(-) &       & 189076.40 & 140.02 & 9(-),10(-),12(-) &       & 189710.07 & 26.35 & 9(+),10(+),11(+) \\
          &       &       & 0.1   &       & 190071.53 & 147.13 & 10(-),11(+),12(-) &       & 190623.40 & 41.52 & 9(+),11(+),12(-) &       & 190023.73 & 130.68 & 9(-),10(-),12(-) &       & 190718.33 & 42.59 & 9(+),10(+),11(+) \\
          \hline
     \parbox[t]{2mm}{\multirow{12}{*}{\rotatebox[origin=c]{90}{uncorrelated}}}  & 37686 & 25    & 0.001 &       & 86132.83 & 65.96 & 10(-),11(-),12(-) &       & 86214.63 & 22.27 & 9(+),11(+) &       & 86158.40 & 41.90 & 9(+),10(-),12(-) &       & 86215.40 & 15.00 & 9(+),10(+) \\
          &       &       & 0.01  &       & 86573.93 & 73.75 & 10(-),11(-),12(-) &       & 86645.23 & 14.75 & 9(+),11(+),12(-) &       & 86614.20 & 27.57 & 9(+),10(-),12(-) &       & 86651.50 & 9.52  & 9(+),10(+),11(+) \\
          &       &       & 0.1   &       & 87128.00 & 53.90 & 10(-),11(-),12(-) &       & 87212.10 & 10.41 & 9(+),11(+) &       & 87162.83 & 29.00 & 9(+),10(-),12(-) &       & 87214.00 & 0.00  & 9(+),10(+) \\
          &       & 50    & 0.001 &       & 84039.10 & 31.64 & 10(-),11(-),12(-) &       & 84059.03 & 13.01 & 9(+),11(-),12(-) &       & 84627.23 & 150.14 & 9(+),10(+),12(-) &       & 84737.87 & 8.97  & 9(+),10(+),11(+) \\
          &       &       & 0.01  &       & 84863.40 & 27.71 & 10(-),11(+),12(-) &       & 84892.63 & 7.00  & 9(+),11(+) &       & 84692.00 & 86.52 & 9(-),10(-),12(-) &       & 84895.63 & 3.30  & 9(+),10(+) \\
          &       &       & 0.1   &       & 85929.30 & 29.21 & 10(-),11(+),12(-) &       & 85958.03 & 15.21 & 9(+),11(+),12(-) &       & 85859.10 & 56.61 & 9(-),10(-),12(-) &       & 85964.77 & 6.08  & 9(+),10(+),11(+) \\
          & 93559 & 25    & 0.001 &       & 147795.97 & 41.29 & 10(-),11(-),12(-) &       & 147878.17 & 8.42  & 9(+),11(+) &       & 147810.67 & 29.52 & 9(+),10(-),12(-) &       & 147875.37 & 9.02  & 9(+),10(+) \\
          &       &       & 0.01  &       & 148168.07 & 41.88 & 10(-),11(-),12(-) &       & 148243.57 & 9.34  & 9(+),11(+),12(-) &       & 148183.73 & 31.21 & 9(+),10(-),12(-) &       & 148247.07 & 8.31  & 9(+),10(+),11(+) \\
          &       &       & 0.1   &       & 148640.40 & 39.29 & 10(-),11(-),12(-) &       & 148714.30 & 9.19  & 9(+),11(+) &       & 148674.13 & 21.49 & 9(+),10(-),12(-) &       & 148713.33 & 7.75  & 9(+),10(+) \\
          &       & 50    & 0.001 &       & 145879.27 & 37.37 & 10(-),11(+),12(-) &       & 145959.83 & 9.18  & 9(+),11(+),12(-) &       & 145660.67 & 77.82 & 9(-),10(-),12(-) &       & 145975.67 & 13.97 & 9(+),10(+),11(+) \\
          &       &       & 0.01  &       & 146609.63 & 51.45 & 10(-),11(+),12(-) &       & 146695.10 & 10.20 & 9(+),11(+) &       & 146508.33 & 43.19 & 9(-),10(-),12(-) &       & 146694.47 & 13.75 & 9(+),10(+) \\
          &       &       & 0.1   &       & 147563.63 & 34.45 & 10(-),11(+),12(-) &       & 147623.67 & 9.87  & 9(+),11(+),12(-) &       & 147529.50 & 39.26 & 9(-),10(-),12(-) &       & 147628.07 & 5.29  & 9(+),10(+),11(+) \\
          \hline
    \end{tabular}}}%
  \label{tab:nagachernoff}%
\end{sidewaystable*}%

\begin{sidewaystable*}[h]
  \centering
  \caption{Statistic results of NSGA-II with Chebyshev's inequality for the instance eil101 with 500 items}
  \scalebox{0.7}{
  \makebox[\linewidth][c]{
  \tabcolsep=0.05cm
    \begin{tabular}{crrrrrrlrrrlrrrlrrrl}
    \hline
          & \multicolumn{1}{l}{capacity} & \multicolumn{1}{l}{delta} & \multicolumn{1}{l}{alpha} &       & \multicolumn{3}{c}{NSGA-II with old and uniform (9)} &       & \multicolumn{3}{c}{NSGA-II with old and PS (10)} &       & \multicolumn{3}{c}{NSGA-II with new and uniform (11)} &       & \multicolumn{3}{c}{NSGA-II with new and PS (12)} \\
          &       &       &       &       & \multicolumn{1}{l}{Mean} & \multicolumn{1}{l}{Std} & stat  &       & \multicolumn{1}{l}{Mean} & \multicolumn{1}{l}{Std} & stat  &       & \multicolumn{1}{l}{Mean} & \multicolumn{1}{l}{Std} & stat  &       & \multicolumn{1}{l}{Mean} & \multicolumn{1}{l}{Std} & stat \\
          \hline
   \parbox[t]{2mm}{\multirow{12}{*}{\rotatebox[origin=c]{90}{bounded-strongly-correlated}}} & 61447 & 25    & 0.001 &       & 73833.80 & 142.46 & 10(-),11(+),12(-) &       & 74461.90 & 22.06 & 9(+),11(+),12(+) &       & 73566.07 & 214.57 & 9(-),10(-),12(-) &       & 74440.57 & 45.01 & 9(+),10(-),11(+) \\
          &       &       & 0.01  &       & 77611.13 & 85.06 & 10(-),11(+),12(-) &       & 77909.53 & 44.31 & 9(+),11(+),12(+) &       & 77293.37 & 159.75 & 9(-),10(-),12(-) &       & 77875.37 & 47.50 & 9(+),10(-),11(+) \\
          &       &       & 0.1   &       & 78663.90 & 145.51 & 10(-),11(+),12(-) &       & 79019.50 & 19.01 & 9(+),11(+),12(+) &       & 78390.00 & 173.59 & 9(-),10(-),12(-) &       & 78998.17 & 40.33 & 9(+),10(-),11(+) \\
          &       & 50    & 0.001 &       & 69066.73 & 163.28 & 10(-),11(+),12(-) &       & 69671.93 & 61.49 & 9(+),11(+) &       & 68813.60 & 167.88 & 9(-),10(-),12(-) &       & 69673.10 & 32.07 & 9(+),11(+) \\
          &       &       & 0.01  &       & 75892.57 & 96.68 & 10(-),12(-) &       & 76304.20 & 29.36 & 9(+),11(+),12(+) &       & 75883.23 & 129.05 & 10(-),12(-) &       & 76275.37 & 19.69 & 9(+),10(-),11(+) \\
          &       &       & 0.1   &       & 78247.07 & 97.79 & 10(-),11(+),12(-) &       & 78507.90 & 4.27  & 9(+),11(+) &       & 78178.97 & 174.67 & 9(-),10(-),12(-) &       & 78506.53 & 5.20  & 9(+),11(+) \\
          & 162943 & 25    & 0.001 &       & 185031.00 & 256.80 & 10(-),11(+),12(-) &       & 186597.97 & 40.87 & 9(+),11(+),12(+) &       & 184857.03 & 229.71 & 9(-),10(-),12(-) &       & 186464.03 & 91.39 & 9(+),10(-),11(+) \\
          &       &       & 0.01  &       & 189960.17 & 161.64 & 10(-),11(+),12(-) &       & 190779.23 & 41.14 & 9(+),11(+),12(+) &       & 189654.13 & 182.06 & 9(-),10(-),12(-) &       & 190732.63 & 58.43 & 9(+),10(-),11(+) \\
          &       &       & 0.1   &       & 191284.70 & 173.66 & 10(-),11(+),12(-) &       & 192124.63 & 73.61 & 9(+),11(+),12(+) &       & 190988.60 & 243.98 & 9(-),10(-),12(-) &       & 192108.20 & 66.57 & 9(+),10(-),11(+) \\
          &       & 50    & 0.001 &       & 178525.63 & 357.61 & 10(-),11(-),12(-) &       & 180523.90 & 45.63 & 9(+),11(+) &       & 178875.80 & 248.68 & 9(+),10(-),12(-) &       & 180519.93 & 46.64 & 9(+),11(+) \\
          &       &       & 0.01  &       & 187701.30 & 165.20 & 10(-),11(-),12(-) &       & 188848.53 & 51.18 & 9(+),11(+) &       & 187879.77 & 169.79 & 9(+),10(-),12(-) &       & 188852.90 & 41.61 & 9(+),11(+) \\
          &       &       & 0.1   &       & 190806.63 & 160.38 & 10(-),11(+),12(-) &       & 191553.40 & 30.92 & 9(+),11(+) &       & 190794.07 & 155.19 & 9(-),10(-),12(-) &       & 191554.83 & 22.00 & 9(+),11(+) \\
          \hline
    \parbox[t]{2mm}{\multirow{12}{*}{\rotatebox[origin=c]{90}{uncorrelated}}} & 37686 & 25    & 0.001 &       & 81097.67 & 120.79 & 10(-),11(+),12(-) &       & 81438.43 & 29.77 & 9(+),11(+),12(-) &       & 79826.57 & 314.75 & 9(-),10(-),12(-) &       & 81457.00 & 0.00  & 9(+),10(-),11(+) \\
          &       &       & 0.01  &       & 86056.53 & 59.24 & 10(-),11(+),12(-) &       & 86172.97 & 18.39 & 9(+),11(+),12(-) &       & 85932.27 & 120.74 & 9(-),10(-),12(-) &       & 86178.23 & 26.97 & 9(+),10(-),11(+) \\
          &       &       & 0.1   &       & 87692.13 & 32.70 & 10(-),11(+),12(-) &       & 87718.70 & 11.17 & 9(+),11(+),12(-) &       & 87666.30 & 59.59 & 9(-),10(-),12(-) &       & 87725.37 & 8.55  & 9(+),10(-),11(+) \\
          &       & 50    & 0.001 &       & 74116.47 & 201.70 & 10(-),11(+),12(-) &       & 74763.10 & 45.28 & 9(+),11(+),12(+) &       & 71446.10 & 514.91 & 9(-),10(-),12(-) &       & 74759.37 & 10.02 & 9(+),10(+),11(+) \\
          &       &       & 0.01  &       & 83765.50 & 102.91 & 10(-),11(+),12(-) &       & 83983.30 & 23.59 & 9(+),11(+) &       & 83404.57 & 514.91 & 9(-),10(-),12(-) &       & 83987.43 & 22.76 & 9(+),11(+) \\
          &       &       & 0.1   &       & 86993.97 & 43.19 & 10(-),11(-),12(-) &       & 87057.27 & 17.19 & 9(+),11(+),12(-) &       & 87024.87 & 35.61 & 9(+),10(-),12(-) &       & 87062.53 & 5.07  & 9(+),10(+),11(+) \\
          & 93559 & 25    & 0.001 &       & 142887.10 & 161.18 & 10(-),11(+),12(-) &       & 143493.53 & 33.13 & 9(+),11(+),12(-) &       & 141780.10 & 281.11 & 9(-),10(-),12(-) &       & 143521.00 & 17.66 & 9(+),10(+),11(+) \\
          &       &       & 0.01  &       & 147612.27 & 78.26 & 10(-),11(-),12(-) &       & 147799.43 & 16.74 & 9(+),11(+) &       & 147602.77 & 89.57 & 9(+),10(-),12(-) &       & 147802.23 & 13.51 & 9(+),11(+) \\
          &       &       & 0.1   &       & 149125.07 & 33.53 & 10(-),11(+),12(-) &       & 149197.40 & 16.73 & 9(+),11(+) &       & 149102.27 & 45.91 & 9(-),10(-),12(-) &       & 149200.67 & 14.26 & 9(+),11(+) \\
          &       & 50    & 0.001 &       & 136249.50 & 232.14 & 10(-),11(+),12(-) &       & 137380.00 & 53.81 & 9(+),11(+),12(-) &       & 133715.83 & 452.04 & 9(-),10(-),12(-) &       & 137442.47 & 26.81 & 9(+),10(+),11(+) \\
          &       &       & 0.01  &       & 144655.87 & 33.53 & 10(-),11(-),12(-) &       & 145808.87 & 18.50 & 9(+),11(+),12(-) &       & 145025.17 & 193.57 & 9(+),10(-),12(-) &       & 145822.27 & 19.88 & 9(+),10(+),11(+) \\
          &       &       & 0.1   &       & 148484.37 & 45.83 & 10(-),11(-),12(-) &       & 148588.10 & 18.61 & 9(+),11(+),12(-) &       & 148517.47 & 38.88 & 9(+),10(-),12(-) &       & 148609.63 & 8.25  & 9(+),10(+),11(+) \\
          \hline
    \end{tabular}}}%
  \label{tab:nsgachebyshev}%
\end{sidewaystable*}%

To compare the performance of the old model with the new model, we first apply the GSEMO to solve the same instances as those proposed in Section \ref{expset}, but with different fitness functions. Next, to test the effectiveness of the PS crossover operator in the multi-objective algorithm, we combined the PS crossover operators in GSEMO to solve the new model. Table \ref{tab:gsemoChernoff} and \ref{tab:gsemoChebyshev} show the results obtained using different probability tails separately. To simplify the algorithm names in the tables, we use \textit{old} denotes the old multi-objective model and \textit{new} is the new multi-objective model. The \textit{uniform}, \textit{HT} and \textit{PS} denote the uniform crossover operator, heavy-tail mutation operator and problem-specific crossover operator separately.

Table \ref{tab:gsemoChernoff} shows that in some of the instances, for example, for the first instance in the \textit{bounded-strongly-correlated} type, GSEMO with the new model outperforms GSEMO with the old model. For the other instances, both algorithms perform as good as each other. It can be observed more clearly in Table \ref{tab:gsemoChebyshev}. The new model performs significantly better than the old model when dealing with the \textit{bounded-strongly-correlated} type of instances. Furthermore, as can be seen from the column \textit{GSEMO with the new model and PS}, this algorithm reported significantly better results than the other two algorithms. These results indicate that the new multi-objective model is effective for solving CCKP instance and performs better than the old model in most cases. The PS crossover operator can improve the performance of the multi-objective evolutionary algorithm when dealing with the CCKP.

\subsection{Experimental Results for NSGA-II}

In this section, we investigate the performance of NSGA-II with the combination of two crossover operators, the uniform crossover and PS crossover, and the two models shown in Section~\ref{multimodel}. We modify NSGA-II to keep the best feasible solution for CCKP in each iteration. Since the NSGA-II generates ten offspring in each step with the population size 20, and GSEMO only generates one offspring, we set the iteration of NSGA-II to $5*10^5$ (instead of $5*10^6$ for GSEMO) which results in the same number of fitness evaluations for both algorithms.

Table \ref{tab:nagachernoff} and \ref{tab:nsgachebyshev} show the results obtained when using Chernoff bound and Chebyshev's inequality, respectively. A significant improvement can be observed from the results obtained for the PS crossover when compared to the uniform crossover.

We now compare the performance of the old model with the new model in a same algorithm. By considering the values in the \textit{stat} columns of Table \ref{tab:nagachernoff}, it can be seen that with the uniform crossover operator, the solutions in the new model \textit{NSGA-II with new and uniform (11)} are mostly better than the solutions in the old model \textit{NSGA-II with old and uniform (9)}. However, in some instances with particular $\delta=50$, the old model outperforms the new model. The solutions in \textit{NSGA-II with old and PS (10)} and \textit{NSGA-II with new and PS (12)} indicate that in most instances, the new model performs better than the old model when using the PS crossover operator. 

However, an opposite conclusion can be drawn from Table \ref{tab:nsgachebyshev}. The solutions in the \textit{NSGA-II with new and uniform (11)} and \textit{NSGA-II with old and uniform (9)} show that NSGA-II with uniform crossover performs better when dealing with the old model than the new model. However, the correlation between the old model and the new model is interesting when using NSGA-II with PS crossover operator to solve the problem. The relationship between the results obtained from solving the two models is related to the type of instances. In other words, for the \textit{bounded-strongly-correlated} instances, the old model outperforms the new model, while for the \textit{uncorrelated} instances, the new model is better than the old model in most cases. 

The next insight can be drawn from the values in the \textit{GSEMO with new and PS (8)} and \textit{NSGA-II with new and PS (12)} columns. It can be observed that the results obtained from GSEMO are significantly better than NGSA-II for all instances. The comparison can point out the possible research line to further investigate state-of-art multi-objective evolutionary algorithms such as NSGA-II and SPEA2 for solving CCKP. Moreover, we compare the performance of the best single-objective algorithm: \textit{$(\mu+1)$~EA with HT and PS} and the best multi-objective algorithm \textit{GSEMO with new and PS} according to the estimated methods. It is observed that the performance of the multi-objective algorithm is significantly better than the single-objective algorithm for all instances.  

\section{CONCLUSION}
\label{conclusion}
In this study, we considered the chance-constrained knapsack problem, which is a variant of the classical knapsack problem. The chance-constrained knapsack problem plays a vital role in various real-world applications, and it allows for constraint violation with a small probability. We have considered the chance-constrained knapsack problem and proposed a problem-specific crossover operator and the heavy-tail mutation operator to deal with the CCKP. Our experimental results show that the proposed operators improve the performance of single-objective evolutionary algorithms when solving CCKP instances. Furthermore, we have introduced a new multi-objective model for the CCKP. The experimental results show that combining this new model with the problem-specific crossover operator in GSEMO and NSGA-II leads to significant performance improvements for solving the CCKP. 
\section{Acknowledgements}
This work has been supported by the Australian Research Council through grant DP160102401 and by the South Australian Government through the Research Consortium "Unlocking Complex Resources through Lean Processing".

\bibliographystyle{abbrv}
\bibliography{main}
\end{document}